\documentclass{ecai} 

\usepackage{latexsym}
\usepackage{amssymb}
\usepackage{amsmath}
\usepackage{amsthm}
\usepackage{booktabs}
\usepackage{enumitem}
\usepackage{graphicx}
\usepackage{color}
\usepackage{algorithm}
\usepackage{algorithmic}
\usepackage{multirow}
\usepackage[table,xcdraw]{xcolor}

\newcommand{\BibTeX}{B\kern-.05em{\sc i\kern-.025em b}\kern-.08em\TeX}

\begin{document}


\begin{frontmatter}


\paperid{8210} 


\title{Solving the Min-Max Multiple Traveling Salesmen Problem via Learning-Based Path Generation and Optimal Splitting}


\author[A]{\fnms{Wen}~\snm{Wang}}
\author[A]{\fnms{Xiangchen}~\snm{Wu}}
\author[A]{\fnms{Liang}~\snm{Wang}\thanks{Corresponding Author. Email: wl@nju.edu.cn}}
\author[A]{\fnms{Hao}~\snm{Hu}}
\author[A]{\fnms{Xianping}~\snm{Tao}}
\author[B]{\fnms{Linghao}~\snm{Zhang}}

\address[A]{Nanjing University}
\address[B]{State Grid Sichuan Electric Power Research Institute}


\begin{abstract}
This study addresses the Min-Max Multiple Traveling Salesmen Problem ($m^3$-TSP), which aims to coordinate tours for multiple salesmen such that the length of the longest tour is minimized. 
Due to its NP-hard nature, exact solvers become impractical under the assumption that $P \ne NP$. 
As a result, learning-based approaches have gained traction for their ability to rapidly generate high-quality approximate solutions. 
Among these, two-stage methods combine learning-based components with classical solvers, simplifying the learning objective. 
However, this decoupling often disrupts consistent optimization, potentially degrading solution quality.
To address this issue, we propose a novel two-stage framework named \textbf{Generate-and-Split} (GaS), which integrates reinforcement learning (RL) with an optimal splitting algorithm in a joint training process.
The splitting algorithm offers near-linear scalability with respect to the number of cities and guarantees optimal splitting in Euclidean space for any given path. 
To facilitate the joint optimization of the RL component with the algorithm, we adopt an LSTM-enhanced model architecture to address partial observability.
Extensive experiments show that the proposed GaS framework significantly outperforms existing learning-based approaches in both solution quality and transferability.
\end{abstract}

\end{frontmatter}


\section{Introduction}

The multiple traveling salesmen problem (mTSP) generalizes the classical traveling salesman problem (TSP) by involving multiple salesmen who depart from a common depot, collectively visit all cities exactly once, and return to the depot.
The mTSP is a versatile model applicable to a wide range of real-world problems \cite{Sun2019MultirobotPP, Chakraa2023AMM, Gan2022ForestFF}.
Solving the mTSP requires generating valid tours for all salesmen while optimizing a specific objective. Two common objectives are the min-sum, which minimizes the total tour length, and the min-max, which minimizes the length of the longest individual tour.
We study the min-max mTSP, abbreviated as $m^3$-TSP, which focuses on improving load balance and reducing worst-case latency.

There lack efficient, polynomial-time solutions to  $m^3$-TSP because the problem is NP-hard \cite{Frana1995TheMS}, and as long as $P \ne NP$.
Classic solvers, based on mathematical programming \cite{cplex2009v12} or heuristic search \cite{Helsgaun2017AnEO}, typically require a substantial amount of time to find a feasible solution. 
In recent years, RL-based methods \cite{Kool2018AttentionLT} have received much research interest because they are promising in generating high-quality solutions efficiently.
One series of RL-based methods, such as DAN \cite{Cao2021DANDA}, ScheduleNet \cite{Park2023LearnTS}, Equity-Transformer (Eqt for short) \cite{Son2023EquityTransformerSN}, and Dpn \cite{Dpn} perform end-to-end training by directly generating solutions to the original $m^3$-TSP and optimizing the process.
Although end-to-end training ensures the searching direction for the optimal solution to the original problem, a reasonable problem decomposition, allowing RL to focus on a simpler part of the problem, is expected to reduce the learning complexity.

Unlike the above work, the other series of work propose to decompose the problem and adopt RL as a part of their solutions.
These methods heuristically decompose the $m^3$-TSP into two interdependent subproblems:
1) \textit{city-division}: dividing the cities into groups, each designated for a specific salesman; and
2) \textit{path-generation}: generating a tour for each salesman to visit the cities within their assigned group.
Following this framework, DisPN \cite{Hu2020ARL} employs RL to first partition the cities among the salesmen, then determines the optimal tour for each salesman by solving smaller-scale TSP instances with classic solvers. 
Differently, SplitNet \cite{Liang2023SplitNetAR} first uses classic solvers to generate an tour visiting all cities (by solving the TSP for a single salesman) and then learns to split this tour for the salesmen.
By decomposing the problem, these approaches effectively simplify the learning process by focusing on city division and leaving path generation to other methods.
However, solving the subproblems independently can lead to sub-optimal overall solutions compared to end-to-end methods.

Inspired by both lines of research, we propose the Generate-and-Split (GaS) framework, an RL-based approach that integrates problem decomposition with joint training to solve the $m^3$-TSP.

Firstly, given an instance of $m^3$-TSP, we train an RL-based \textbf{path generator} that generates a permutation of cities step-by-step, with each step selecting an unvisited city.
Unlike end-to-end methods that address city division and path generation simultaneously by creating specific routes for each salesman, the path generator finds an path visiting all cities, which is then split into tours and assigned to individual salesmen---similar to SplitNet's approach.
By focusing on learning for path generation, we simplify the problem, reducing it to one with a shorter decision horizon.
In contrast to prior approaches such as DisPN and SplitNet, which address subproblems in isolation, the learning-based generator operates within a unified training process, aiming to produce paths that facilitate low-cost solutions through the utilization of the splitting algorithm.
Secondly, we present a straightforward and efficient, optimal \textbf{splitting algorithm} designed to split any specified path into tours.
The splitting algorithm divides a sequence of length $n-1$ into $m$ subsequences, with an objective of minimizing the maximum cost of the tours corresponding to the subsequences, where $n-1$ refers to the number of cities (excluding the depot), and $m$ is the number of traveling salesmen.
A brute-force method can solve the splitting problem with a prohibitive time complexity of $O(n^{m-1})$ by enumerating all possible split points.
As a result, SplitNet adopts a learning-based approach to find effective splitting strategies in a computationally efficient manner.
Unlike SplitNet, in this work, we show that \textit{we can find optimal (without considering the float-pointing-error) splitting strategies with an efficient, deterministic algorithm, given that we define $m^3$-TSP in the Euclidean space}.
More specifically, we propose a splitting algorithm that can approximate the optimal solution with an approximation ratio of $1+\epsilon$, $\forall \epsilon>0$, with a time complexity of $O(n \log \frac{\hat{c}}{\epsilon})$ for any given path.
$\hat{c}$ is the maximum possible cost, and $\epsilon$ represents the desired precision. 
Thirdly, since the splitting procedure is implemented subsequent to generating the entire path, essential information—such as the number of traveling salesmen who have finished their tours—is not accessible at the moment decisions are made.
This situates the challenges of partial observation in the framework of RL.
To address this, we model the task as a partially observable Markov decision process (POMDP) \cite{Kaelbling1998PlanningAA} and adopt an LSTM-based architecture \cite{Hochreiter1997LongSM} to mitigate the effects of partial observability.
The use of recurrent neural networks (RNNs) to tackle partial observability dates back to the early development of deep RL \cite{Hausknecht2015DeepRQ}, and has since been adopted in combinatorial optimization frameworks \cite{bogyrbayeva2022deepreinforcementlearningapproach, pmlr-v222-iklassov24a}, an idea we also draw upon in this work.

In summary, this work makes the following contributions: 

\begin{itemize} 
    \item We propose the GaS framework for the $m^3$-TSP problem, which adopts the principle of problem decomposition. 
    It integrates a path generator to tackle the more challenging aspects of the problem, and a deterministic splitting algorithm to efficiently address the remaining components.
    \item We design an LSTM-enhanced decoder to support joint learning together with the splitting algorithm. The splitting algorithm, though simple and efficient, plays a crucial role by providing effective reward signals for the RL-based path generator.
    \item We conduct extensive experiments to evaluate the solution quality and transferability of the proposed GaS framework. The results show that, compared to strong baselines such as Eqt \cite{Son2023EquityTransformerSN} and Dpn \cite{Dpn}, our approach achieves a statistically significant performance improvement ($p < 0.05$) and demonstrates better transferability in 90.7\% of the out-of-distribution data.
\end{itemize}

\begin{figure*}[!htbp]
    \includegraphics[width=\textwidth]{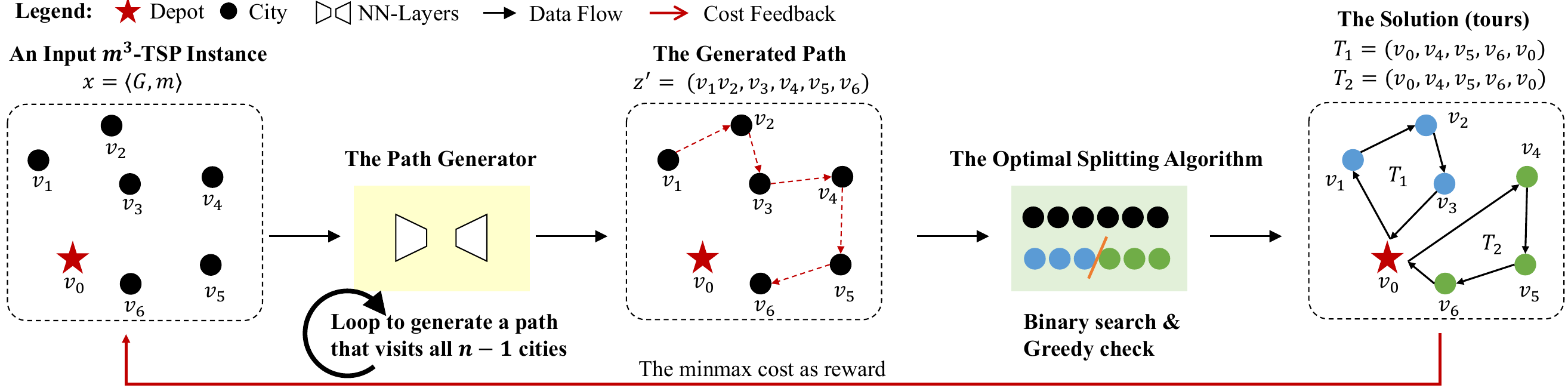}
    \caption[short]{The GaS framework. 
    The two key components are the path generator and the optimal splitting algorithm.
    }
    \label{fig:gas_framework}
\end{figure*}

\section{Problem Formulation}\label{sec:formulation}

The $m^3$-TSP is defined on a \textit{complete} graph $G=(V, E)$, with $V = \{v_0, v_1, \cdots, v_{n-1}\}$ denotes the set of $n$ nodes (which include a depot $v_0$, and $n-1$ cities from $v_1$ to $v_{n-1}$), and $E$ denotes the set of edges.
Following existing work \cite{Liang2023SplitNetAR}, we place the nodes on a 2-dimensional Euclidean plane with $(x_{v}, y_{v})$ denotes the coordinate of a node $v\in V$.
Each edge has a weight $d(v_i, v_j)$ corresponding to the Euclidean distance between nodes $v_i$, and $v_j$.

There are $m$ salesmen (agents) who start their travel from the depot to visit the cities before returning to the depot.
The tour of the $i$-th agent is defined as a sequence of nodes $T_i = (v_{i, 0}, v_{i, 1}, v_{i, 2}, \cdots, v_{i, n_i}, v_{i, n_i + 1})$, with $v_{i, 0} = v_{i, n_i + 1} = v_{0}$ being the depot.
The tour will not contain any repeated nodes except for the depot, and the depot node will not appear in the middle of the tour,
meaning that $\forall{1 \le j, k \le n_i}, v_{i, j} \ne v_{i, k} \ne v_0$.
We use $\{T_i\}$ to denote the set of nodes in tour $T_i, i=1, 2, \cdots, m$, with $|T_i| = n_i + 2$ being the number of nodes in $T_i$.
Every node except the depot must be visited exactly once by the agents, i.e., we have $\cup_{i=1}^{m} \{T_i\} = V$, and $\cap_{i=1}^{m} \{T_i\} \setminus \{v_0\} = \emptyset$.
The cost of the $i$-th agent's tour is defined in equation (\ref{equ:mtsp_cost}).

\begin{equation}\label{equ:mtsp_cost}
    C(T_i) = \sum_{j=1}^{|T_{i}| - 1} d(v_{i, j-1}, v_{i, j}).   
\end{equation}

The objective of solving the $m^3$-TSP is to plan tours for all agents such that the maximum cost among the agents' tours is minimized, which is defined in equation (\ref{equ:minmax_objective}).

\begin{equation}\label{equ:minmax_objective}
\begin{aligned}
    L(z^*)  &= \min \max \{ C(T_i)|, 1 \le i \le m \} \\
    s.t. &\cup_{i=1}^{m} \{T_i\} = V \\
    &\cap_{i=1}^{m} \{T_i\} \setminus \{v_0\} = \emptyset \\
    &\forall_{1 \le i \le m, 1 \le j, k \le n_i} T_{i, j} \ne T_{i, k} \ne v_0
\end{aligned}
\end{equation}

\noindent where $L(z^*)$ represents the optimal cost for the $m^3$-TSP, and $z^*$ is the optimal solution for $m^3$-TSP.
In this context, $z^*$ is defined as $T_1 + T_2 + \cdots + T_m$, where the $+$ symbol representing sequence concatenation.

\section{Method}\label{sec:method}

In this section, we introduce the GaS framework, as illustrated in Figure~\ref{fig:gas_framework}.
We propose to first generate a path that sequentially visits all cities using an RL-based path generator and split the path to form the tours for each salesman with a splitting algorithm.
The path generator employs RL to output the sequence of all cities, excluding the depot. 
Then we split the sequence into multiple subsequences by the splitting algorithm, which are subsequently assigned to the corresponding traveling salesmen. 
After the assignment is complete, the maximum cost of the tours corresponding to each subsequence is used as the reward to train the path generator.

The meta-level idea behind the design of this framework is problem decomposition. 
The $m^3$-TSP can be naturally divided into two components: a navigation component (TSP-like) and a partitioning component \cite{Dpn}. 
While solving the TSP is NP-complete \cite{Papadimitriou1977TheET}, the partitioning part does not necessarily pose significant computational difficulty, provided an appropriate problem decomposition is adopted.
For the components that are indeed hard to solve, we rely on RL, which motivates the design of the path generator. 
For the components that may admit polynomial-time algorithms, we design the splitting algorithm to ensure solution quality.
The idea of decomposition not only enhances the interpretability and modularity of the framework, but also offers potential for integration with a broader spectrum of learning-based approaches.
In the following subsections, we introduce the path generator and the splitting algorithm in detail.

\subsection{The Path Generator}

Given an instance of the $m^3$-TSP, \textbf{the path generator} is responsible for generating a path across all the cities:

\begin{equation}
    z' = (v_{1'}, v_{2'}, \cdots, v_{i'}, \cdots, v_{n-1'})
\end{equation}
\noindent where $v_{i'}, 1\leq i' \leq n-1$ is a city node defined in the graph.
The path will be split and assigned to different salesmen, by the \textbf{splitting algorithm}, as detailed in the following section. 

We formulate the problem of generating a path given an instance of $m^3$-TSP using the partially observable Markov Decision Processes (POMDPs) and train the path generator that generates a path step-by-step, similar to AM \cite{Kool2018AttentionLT} and Eqt \cite{Son2023EquityTransformerSN}, as follows.

\subsubsection{POMDP Formulation}

\textbf{Observation.} We define the observation of the problem at time $t$ as $\textbf{s}_t = \langle \textbf{x}_g, \textbf{o}_t \rangle$, where $\textbf{x}_g = \langle G, k\rangle$ contains time-independent global information, and $\textbf{o}_t = \langle n_f, n_c\rangle$ contains observations made at time $t$.
In $\textbf{x}_g$, $G$ represents the complete graph that defines an instance of the problem as introduced in the previous section, and $k = n/m$ is the ratio of the number of cities to the number of salesmen.
Here, $G$ is represented by the 2D coordinates of all cities and the depot.
In $\textbf{o}_t$, $n_f$ represents the information of the depot, and $n_c$ represents the information of the currently visited city.

From the definition of the observation, it is clear that key information (state) necessary for solving the problem, such as the number of traveling salesmen currently in use, is not included. This omission arises because our framework delegates the decision of when to return to the depot to the splitting algorithm, making such information unavailable during the policy execution. Unlike work such as Eqt \cite{Son2023EquityTransformerSN}, our observation definition adheres to the principle of simplicity, retaining only the essential information that can be obtained, without introducing additional heuristics.
The parts that are difficult to define explicitly are instead learned by the LSTM.

\textbf{Action.} The action $a_t$ can be any unvisited city at time $t$.
Notably, the depot is not considered a city, so the action sequence will exclude the depot.
After all cities have been visited, the action sequence forms $z'$, which is the result of the path generator.

\textbf{Reward.} The reward is the negative cost of the maximum of $C(T_i)$ as described in equation (\ref{equ:mtsp_cost}) at the final time step.
At all other steps, the reward is zero.
To obtain a reward from $z'$, it must first be split by the splitting algorithm described in the next section.

\subsubsection{Policy Description}

When addressing the problem using RL, we aim to learn a policy $\pi_\theta(a_t \mid \mathbf{s}_{1:t})$, which defines a probability distribution over the action space conditioned on the sequence of past observations. The objective is to maximize the expected cumulative reward, which, in our setting, corresponds to the optimization goal of the $m^3$-TSP. The probability distribution over a complete solution path is denoted as $p_\theta(z' | x)$, where $x$ denotes a specific $m^3$-TSP instance, $z'$ is a candidate solution, and $\theta$ represents the learnable parameters of the policy, as detailed in the next section.

\subsection{Network Architecture of the Path Generator}

\begin{figure}[!htbp]
    \includegraphics[width=\columnwidth]{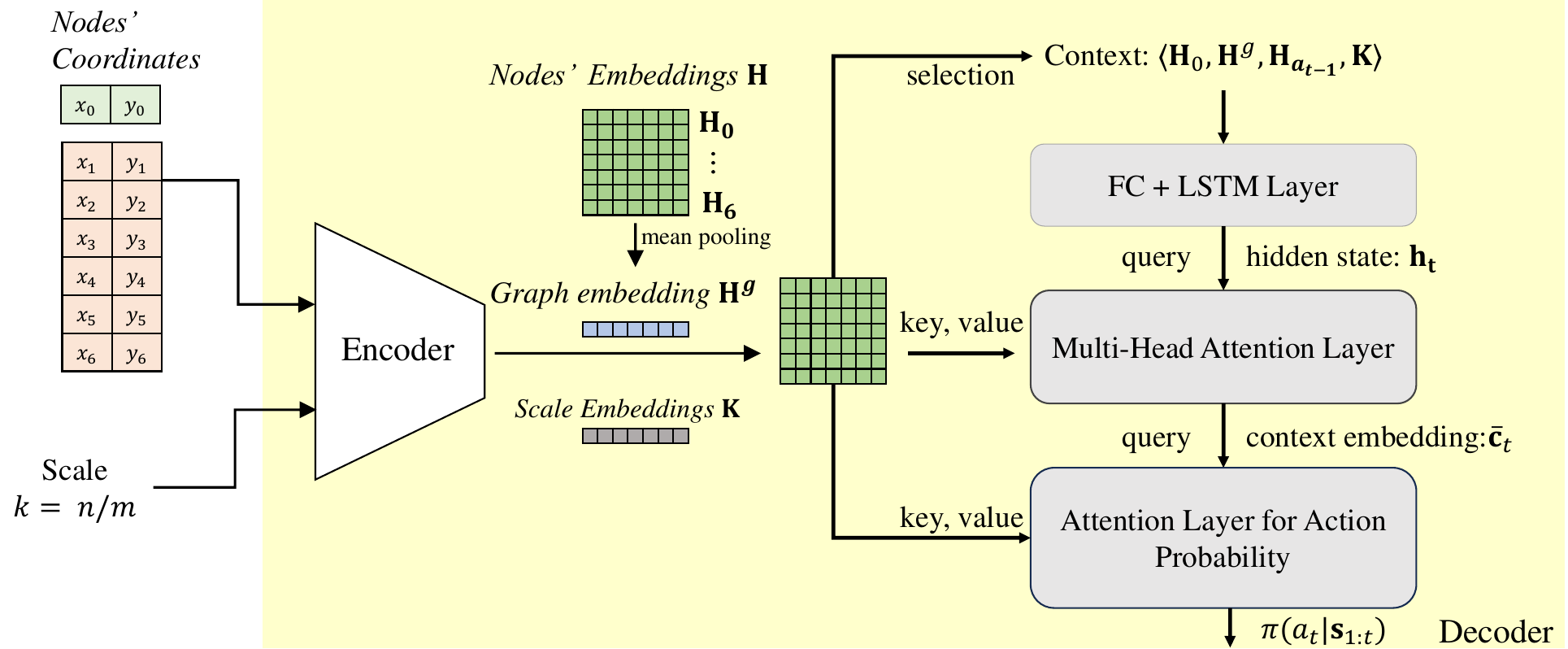}
    \caption[short]{The network architecture of the path generator.}
    \label{fig:network}
\end{figure}

The splitting algorithm can only be applied after the entire path is generated, which presents a challenge for training the path generator by RL, known as the partial observation problem due to delayed splitting.
From the definition of the POMDP, we observe that at each time step, the state should include information about which salesman is taking action at this time step.
Methods like Eqt \cite{Son2023EquityTransformerSN} include the ``return to the depot'' in the action space, allowing the environment to naturally switch to the next traveling salesman and update state transitions accordingly.

To tackle this issue, inspired by the RL for POMDPs, we introduce an LSTM-based structure in the decoder that efficiently estimates the true state under the current policy execution history for a given $m^3$-TSP instance.
This structure constructs a hidden state based on the observation history and makes subsequent decisions based on the hidden state.
The network structure is shown in Figure \ref{fig:network}.

\subsubsection{Encoder}

The encoder takes as input the two-dimensional coordinate data of the nodes and the total number of traveling salesmen. 
The output includes node embedding, a graph embedding and scale embedding for the $m^3$-TSP instance.
The initial embedding of all nodes is computed by applying a fully connected layer to transform the node coordinates, where distinct transformations are employed for the depot node and the remaining city nodes. Following this, multiple attention layers as described in \cite{Kool2018AttentionLT} are employed to process the node embedding, culminating in the embedding $\mathbf{H}$ for all nodes.
The embedding of the $i$-th node is denoted by $\mathbf{H}_i$. 
To derive the graph embedding $\mathbf{H}^g$, we employ mean pooling \cite{NEURIPS2020_1764183e} across the node embedding. 
The scale embedding $\mathbf{K}$ is acquired by mapping the scale $k = n/m$ of a $m^3$-TSP instance into the embedding space via a fully connected layer.
We employ scale embedding to furnish the model with scale-specific information, thereby enabling it to generalize across various problem scales.

\subsubsection{Decoder}

The decoder takes the nodes' embedding and scale embedding as input and outputs the probability of each legal action being selected at each time step.
Compared to the classical decoder structure, we have added several components to mitigate partial observation problems. 
The relevant concepts are presented as follows.

\textbf{Context}: At the $t$-th step of the path generation, $t-1$ actions have already been generated, denoted as $\langle a_1, a_2, \cdots, a_{t-1} \rangle$.
The context of the problem at this step is defined as the concatenation of vectors: $\mathbf{c}_t = [\mathbf{H}^g ; \textbf{H}_0, \mathbf{H}{a_{t-1}} ; \textbf{K}]$.
The vector $\mathbf{c}_t$ is then passed through a fully connected layer, the result is denoted as $\hat{\mathbf{c}}_t$.

\textbf{Hidden State}: To mitigate the partial observability problem, we employ an LSTM \cite{Hochreiter1997LongSM} to process the context vector $\hat{\mathbf{c}}_t$ and maintain a hidden state $\mathbf{h}_t$.
Specifically, we adopt an incremental formulation, where the hidden state is updated as $\mathbf{h}_t = \mathrm{LSTM}(\hat{\mathbf{c}}_t, \mathbf{h}_{t-1})$, with the initial states $\mathbf{h}_0$ set to zero vectors.

A new context embedding from hidden state is computed by a multi-head attention layer $\overline{\mathbf{c}}_t = \mathbf{MHA}(\mathbf{h}_t, \mathbf{H}, \mathbf{H})$ with appropriate mask. Finally, $\overline{\mathbf{c}}_t$ is used as a contextual query, with all valid nodes' embedding serving as keys and values, to construct an attention-based policy $\pi_{\theta}(a_t|\mathbf{s}_{1:t})$, consistent with the AM \cite{Kool2018AttentionLT}. 

\subsection{The Splitting Algorithm}

\renewcommand{\algorithmicrequire}{\textbf{Input:}}
\renewcommand{\algorithmicensure}{\textbf{Output:}}

\renewcommand{\algorithmicrequire}{\textbf{Input:}}
\renewcommand{\algorithmicensure}{\textbf{Output:}}
\begin{algorithm}[!t]
    \caption{Greedy Check}\label{alg:greedy_check}
    \begin{algorithmic}[1] 
        \REQUIRE $G$, graph consisting of a depot and cities; 
        \REQUIRE $m$, the number of agents; 
        \REQUIRE $z'$, path generated by the path generator;
        \REQUIRE $c_m$, the threshold to check.
        \ENSURE whether the threshold can be successfully achieved.
        \STATE $i \gets 0$
        \STATE initialize $T_i$ as $(v_0, )$

        \FOR {$t = 1, 2, \cdots, n - 1$}
            \IF {$C(T_i + z'_t + v_0) \le c_m$ // equation (\ref{equ:mtsp_cost})} 
                \STATE $T_i \gets T_i + z'_t$
            \ELSE
                \STATE $T_{i+1} = (v_0, z'_t, )$
                \IF {$i + 1 \ge m$ or $C(T_{i+1} + v_0) > c_m$}
                    \RETURN \FALSE
                \ENDIF
                \STATE $i \gets i + 1$
            \ENDIF
        \ENDFOR
        \RETURN \TRUE
    \end{algorithmic}
\end{algorithm}

\begin{algorithm}[!t]
    \caption{The Splitting Algorithm}\label{alg:bsp}
    \begin{algorithmic}[1] 
        \REQUIRE $G$, graph consisting of a depot and cities; 
        \REQUIRE $m$, the number of agents; 
        \REQUIRE $z'$, path generated by the path generator. 
        \ENSURE $c_m$ is the near optimal cost.
        
        \STATE $L = 0, R = \hat{c}$ // binary search boundary

        \WHILE {$R - L > \epsilon$}
            \STATE $c_m \gets (L + R) / 2$

            \IF {GreedyCheck($G, m, z'$)}
                \STATE $L \gets c_m$
            \ELSE
                \STATE $R \gets c_m$
            \ENDIF
        \ENDWHILE
    \end{algorithmic}
\end{algorithm}

After the path is generated, we need to split it to obtain the solution for the $m^3$-TSP. 
There are various possible splitting schemes, and each results in a different cost. 
Our goal is to efficiently compute the cost corresponding to the optimal splitting scheme, which will be used as the reward for RL training.
Assuming the path generator provides the visiting path $z'$ for all salesmen,
we need to split the path into $m$ subsequences and insert $v_0$ as the first and last elements of the subsequences to form tours $\{T_i|1 \le i \le m\}$ for each salesman.
The min-max cost of the tours can be calculated in the splitting algorithm, as defined by equation (\ref{equ:mtsp_cost}).
A brute-force algorithm can enumerate all possible splitting points to obtain the optimal result with a $O(n^{m-1})$ time complexity.
The time complexity of such an algorithm increases exponentially with the number of salesmen, which is not suitable for fast training.
To address this, we design a splitting algorithm with near-linear time complexity with respect to the number of nodes, which is outlined below.

Firstly, we convert this problem into a decision problem as follows: 
\textit{Given a path $z'$, can it be visited in order by no more than $m$ salesmen such that the tour length of any salesmen does not exceed a predefined threshold $c_m$?}
This decision problem can be addressed using a greedy algorithm, referred to as ``greedy check'' as listed in Algorithm \ref{alg:greedy_check}.
In the algorithm block, we use ``+'' for sequence concatenation.
In this greedy check, each traveling salesman attempts to visit as many cities as possible. 
The answer to the original decision problem is determined by whether more than $m$ salesmen are required by greedy check.
If the greedy check uses more than $m$ salesmen, then no other method can use no more than $m$ salesmen to finish the task, so the answer to the decision problem is ``no''.
Conversely, it provides a solution to the decision problem, so the answer to the decision problem is ``yes''.
Secondly, If the decision problem is satisfied for a given $c_m$, then it will also be satisfied for any $c \ge c_m$.
Otherwise, it cannot be satisfied for any $c \le c_m$.
This suggests that the minimal $c_m$ that satisfies the decision problem can be efficiently determined using binary search.
By synthesizing the aforementioned aspects, we introduce the splitting algorithm detailed in Algorithm \ref{alg:bsp}.
The time complexity of the algorithm is $O(n \log \frac{\hat{c}}{\epsilon})$, and the approximation ratio is $1 + \epsilon$ (or $1 + \epsilon / c^*$ if we assume $c^* < 1$, where $c^*$ is the optimal cost). 
The cost reported by the splitting algorithm is used as a reward for RL, denoted as $L(z')$.

As mentioned earlier, if the greedy check can correctly answer the decision problem, binary search can find the optimal solution within the error margin.
We provide its proof in the Appendix.

\subsection{The Training Process}

RL trains through the REINFORCE \cite{Williams2004SimpleSG}, with the gradient of the objective function given by Equation (\ref{equ:objective_rl}).

\begin{equation}\label{equ:objective_rl}
    \nabla_\theta = {\rm E}_{p_\theta(z'|x) }[\nabla \log p_\theta(z'|x) (L(z') - b(x))]
\end{equation}

In Equations (\ref{equ:objective_rl}), $x$ represents an instance of the $m^3$-TSP, $p_\theta(z'|x)$ denotes the probability of the policy generating the path $z'$ condition on $x$, and $b(x)$ denotes the baseline, following \cite{Son2023EquityTransformerSN}, which is computed as the average reward over multiple trajectories generated by applying data augmentation to the input instance $x$.
Several similar symmetry-based data augmentation techniques can also be found in \cite{yu2024leveraging, yu2023esp, yu2024adaptaug}.
Finally, a gradient descent-based method that can be used to optimize the problem.

\section{Experimental Results}\label{sec:experiments}

In this section, we aim to answer the following research questions:

\begin{enumerate}
    \item RQ1 [Solution Quality]. Is the proposed GaS framework effective and efficient in finding high-quality solutions?
    \item RQ2 [Transferability]. Is a learned solver still effective when we apply it to problems different from the training scenarios in terms of scales and city distributions?
\end{enumerate}

\subsection{Baseline Selection}

Several deep RL-based methods have been proposed for the $m^3$-TSP. Among them, we select two of the most recent and competitive algorithms for comparison: Eqt \cite{Son2023EquityTransformerSN} and Dpn \cite{Dpn}.
Our selection is based on the following criteria: (1) both methods have demonstrated impressive empirical performance and, to the best of our knowledge, represent the most recent and state-of-the-art learning-based approaches; and (2) both are fully open-sourced, enabling fair comparisons under identical experimental settings and ensuring the reproducibility of results.
We directly used the publicly available models of Eqt and Dpn for evaluation. Since Eqt does not provide a pretrained model for instances with 100 nodes, we trained Eqt for this setting using its publicly released training script.

For classical heuristic methods, following the evaluation protocol established in Eqt \cite{Son2023EquityTransformerSN}, we report the performance of LKH3 \cite{Helsgaun2017AnEO} on the $m^3$-TSP.
This allows us to quantify the performance gap between learning-based approaches and traditional heuristic algorithms.
HGA \cite{mahmoudinazlou2024hybrid} is another competitive heuristic solver, but due to space constraints, we restrict our comparisons to LKH3.
We report heuristic solver results under 60 and 600 second time limits.

\subsection{Experiment Design}

We conduct experiments under the following controlled settings.

To address \textbf{RQ1}, we randomly generate 2D coordinates for the cities and depot, uniformly distributed within a $1 \times 1$ rectangular area during training. The RL-based policy is trained under two settings: $n=50$ and $n=100$ nodes. The number of salesmen is randomly sampled from a uniform distribution in the range $[2, 10]$. We then validate the trained GaS framework on 100 randomly generated instances, using the same test data as in Eqt \cite{Son2023EquityTransformerSN} and Dpn \cite{Dpn}. These instances are not part of the training process.

The answer to \textbf{RQ1} focuses on the solution quality of the GaS framework for problems of the same scale and distribution. However, in practical applications, we often encounter large-scale problems \cite{Li2021LearningTD} and out-of-distribution data \cite{zhou2023towards}. Instances of varying scales are also crucial. To assess the broader applicability of the framework, we conduct additional experiments to address \textbf{RQ2}. Specifically, we perform three types of experiments as follows:
\begin{enumerate}
    \item \textbf{RQ2.1}: How does the trained model perform on data with different distributions? To answer this, we directly apply the learned model to three out-of-distribution datasets proposed in \cite{zhou2023towards} and compare the results with baseline methods.
    \item \textbf{RQ2.2}: How does the trained model perform on large-scale problems? To answer this, we employ a similar fine-tuning process on scenarios with 200 and 500 nodes, as described in \cite{Dpn}, and compare the results with baseline methods on large-scale problems.
    \item \textbf{RQ2.3}: How does the trained model perform when the scale differs from the training settings? To answer this, we evaluate the fine-tuned model on $n=200$ and $n=500$ settings with varying numbers of salesmen.
\end{enumerate}

Our code and data are publicly available~\cite{gas2025github}.
For further details regarding reproducibility, please refer to Appendix. 

\subsection{Solution Quality Results (RQ1)}\label{sec:eval_same}

The experimental results for RQ1 are summarized in Table \ref{tab:tab1}.
In the table, $\times k.$ represents the generation of $k$ samples and reporting the best result. 
For Eqt and GaS, $\times 8$ indicates 8-fold data augmentation \cite{Kwon2020POMOPO}, while $\times 128$ refers to performing 16 sampling width \cite{Kool2018AttentionLT} for each instance after 8-fold data augmentation, and reporting the best result. 
On the other hand, Dpn utilizes the data augmentation and agent permutation techniques described in the referenced paper.
As shown in Table \ref{tab:tab1}, the GaS method achieves better results than the best existing learning-based methods in 13 out of 18 experimental configurations, and matches their performance in one other.
One-sided Wilcoxon tests \cite{wilcoxon1945individual} show that GaS significantly reduces the cost  compared to Dpn and Eqt, with $p$-values of $9.66 \times 10^{-29}$ / $1.10 \times 10^{-42}$ (50 nodes) and $2.91 \times 10^{-7}$ / $4.00 \times 10^{-80}$ (100 nodes).
A comparison between learning-based methods and LKH3 reveals that existing learning-based approaches still lag behind search-based methods like LKH3 in terms of solution quality, particularly when the value of $m$ is small. This indicates that when the problem characteristics are closer to the classical TSP, current learning-based methods tend to perform poorly and leave greater room for improvement. In contrast, when $m$ is large, learning-based methods are already capable of solving the problem effectively, leaving less room for further optimization.
We do not report solution time in the table because all learning-based approaches solve each instance in under one second at this scale, making the differences among them negligible. Nevertheless, compared to the search-based LKH3, learning-based methods demonstrate a clear advantage in terms of inference efficiency.

\begin{table*}[!htbp]
\centering
\caption{\textbf{Evaluation Results for RQ1}. The average min-max costs are reported in this table, with results averaged over 100 instances. A gray background highlights the best performance among learning-based methods, while bold font indicates the best performance across all methods.}
\label{tab:tab1}
\resizebox{\textwidth}{!}{%
\begin{tabular*}{\textwidth}{@{\extracolsep{\fill}} cccccccccc}
\toprule
\multicolumn{10}{c}{\textbf{mTSP Instances, n=50}} \\ \midrule
\multicolumn{1}{c|}{m} & 2 & 3 & 4 & 5 & 6 & 7 & 8 & 9 & 10 \\ \midrule
LKH3 (60s) & 3.1520 & 2.4353 & 2.1546 & 2.0340 & 1.9807 & 1.9515 & 1.9481 & 1.9400 & 1.9431 \\
LKH3 (600s) & \textbf{3.1517} & \textbf{2.4338} & \textbf{2.1502} & \textbf{2.0234} & \textbf{1.9711} & \textbf{1.9440} & 1.9358 & 1.9336 & 1.9317 \\ \midrule
Eqt $\times 1$ & 3.3262 & 2.5643 & 2.2402 & 2.0798 & 1.9976 & 1.9618 & 1.9414 & 1.9349 & 1.9321 \\
Eqt $\times 8$ & 3.2343 & 2.4901 & 2.1870 & 2.0399 & 1.9788 & 1.9465 & 1.9359 & 1.9324 & 1.9303 \\
Eqt $\times 128$ & 3.2079 & 2.4700 & 2.1732 & 2.0315 & 1.9742 & 1.9443 & \cellcolor{gray!30} \textbf{1.9349} & \cellcolor{gray!30} \textbf{1.9321} & 1.9303 \\ \midrule
Dpn $\times 1$ & 3.2576 & 2.5149 & 2.2038 & 2.0545 & 1.9979 & 1.9549 & 1.9430 & 1.9352 & 1.9334 \\
Dpn $\times 8$ & 3.1956 & 2.4654 & 2.1735 & 2.0337 & 1.9744 & 1.9460 & 1.9362 & 1.9323 & 1.9304 \\
Dpn $\times 128$ & 3.1949 & 2.4637 & 2.1728 & 2.0323 & \cellcolor{gray!30} 1.9736 & 1.9454 & 1.9358 & 1.9323 & \cellcolor{gray!30} \textbf{1.9302} \\ \midrule
GaS $\times 1$ & 3.3339 & 2.5448 & 2.2216 & 2.0575 & 1.9923 & 1.9572 & 1.9396 & 1.9338 & 1.9312 \\
GaS $\times 8$ & 3.2108 & 2.4723 & 2.1788 & 2.0367 & 1.9771 & 1.9451 & 1.9354 & 1.9323 & 1.9307 \\
GaS $\times 128$ &  \cellcolor{gray!30} 3.1918 & \cellcolor{gray!30} 2.4598 & \cellcolor{gray!30} 2.1681 & \cellcolor{gray!30} 2.0301 & 1.9739 & \cellcolor{gray!30} 1.9440 & \cellcolor{gray!30} \textbf{1.9349} & 1.9322 & 1.9305 \\ \midrule
\multicolumn{10}{c}{\textbf{mTSP Instances, n=100}} \\ \midrule
\multicolumn{1}{c|}{m} & 2 & 3 & 4 & 5 & 6 & 7 & 8 & 9 & 10 \\ \midrule
LKH3 (60s) & 4.0713 & 2.9551 & 2.4814 & 2.2397 & 2.1119 & 2.0580 & 2.0360 & 2.0225 & 2.0090 \\
LKH3 (600s) & \textbf{4.0694} &\textbf{ 2.9436} & \textbf{2.4572} & \textbf{2.2058} & \textbf{2.0719} & \textbf{2.0076} & 1.9799 & 1.9689 & 1.9640 \\ \midrule
Eqt $\times 1$ & 4.5123 & 3.2110 & 2.6303 & 2.3239 & 2.1528 & 2.0557 & 2.0036 & 1.9760 & 1.9634 \\
Eqt $\times 8$ & 4.3148 & 3.1044 & 2.5549 & 2.2646 & 2.1083 & 2.0290 & 1.9845 & 1.9637 & 1.9549 \\
Eqt $\times 128$ & 4.2524 & 3.0672 & 2.5280 & 2.2469 & 2.0947 & 2.0164 & 1.9792 & 1.9611 & 1.9531 \\ \midrule
Dpn $\times 1$ & 4.2705 & 3.0801 & 2.5498 & 2.2704 & 2.1177 & 2.0335 & 1.9921 & 1.9724 & 1.9587 \\
Dpn $\times 8$ & 4.1804 & 3.0233 & 2.5041 & 2.2346 & 2.0909 & 2.0143 & 1.9804 & 1.9623 & 1.9534 \\
Dpn $\times 128$ & \cellcolor{gray!30} 4.1774 & 3.0181 & 2.5015 & 2.2315 & 2.0895 & 2.0128 & 1.9793 & 1.9613 & 1.9531 \\ \midrule
GaS $\times 1$ & 4.3156 & 3.1230 & 2.5766 & 2.2796 & 2.1240 & 2.0429 & 1.9996 & 1.9739 & 1.9584 \\
GaS $\times 8$ & 4.2267 & 3.0439 & 2.5114 & 2.2417 & 2.0964 & 2.0190 & 1.9807 & 1.9625 & 1.9539 \\
GaS $\times 128$ & 4.1852 & \cellcolor{gray!30} 3.0157 & \cellcolor{gray!30} 2.4974 & \cellcolor{gray!30} 2.2285 & \cellcolor{gray!30} 2.0859 & \cellcolor{gray!30} 2.0120 & \cellcolor{gray!30} \textbf{1.9764} & \cellcolor{gray!30} \textbf{1.9596} & \cellcolor{gray!30} \textbf{1.9524} \\ \bottomrule
\end{tabular*}%
}
\end{table*}

\subsection{Transferability Results (RQ2)}

\begin{figure}[!htbp]
	\centering
	\includegraphics[width=0.8 \linewidth]{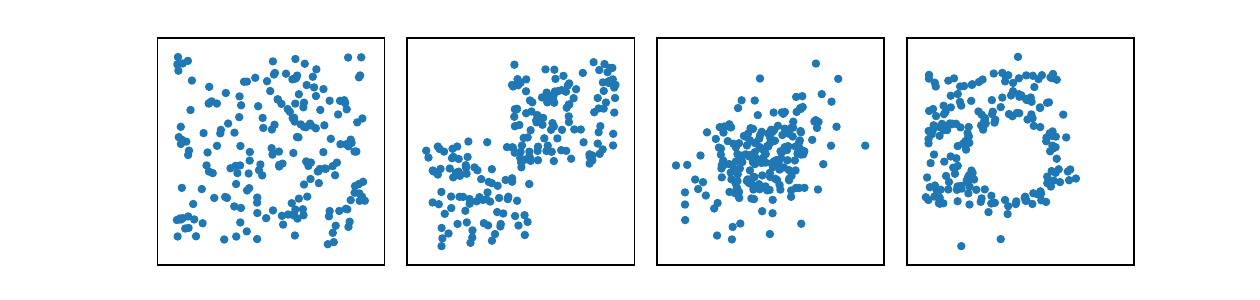}
    \vspace{-15pt}
    \caption{City locations under different distributions. From left to right: Uniform, Rotation, Gaussian, Explosion.}
    \label{fig:distributions}
\end{figure}

\textbf{Results for RQ2.1}: The trained model should be applicable to data with distributions different from the training distribution. We evaluated the model on datasets with Rotation, Gaussian, and Explosion distributions \cite{zhou2023towards}. A sample of the data is shown in Figure \ref{fig:distributions}, and the comparison results with Eqt and Dpn are presented in Table 2. The original dataset contains 200 nodes and 2000 instances for each distribution. For evaluation, we used the first 100 instances and the first $n$ nodes.
According to the experimental results, GaS outperforms existing learning-based methods in 41 out of 54 test configurations and achieves identical performance (to four decimal places) in 8 more. This means that in 90.7\% of the test cases, GaS performs at least as well as the best current learning-based approaches.

\begin{table*}[!htbp]
\centering
\caption{\textbf{Evaluation Results for RQ2.1}. The average min-max cost are reported in this table. All results are averaged over 100 out of distribution instances. A gray background indicates the best performance among learning-based
methods.}
\label{tab:my-table}
\resizebox{\textwidth}{!}{%
\begin{tabular*}{\textwidth}{@{\extracolsep{\fill}} cccccccccc}
\toprule
\multicolumn{1}{c|}{m} & 2               & 3               & 4               & 5               & 6               & 7               & 8               & 9               & 10              \\ \midrule
\multicolumn{10}{c}{\textbf{Rotation, n = 50}}                                                                                                                                           \\ \midrule
Eqt $\times 128$ & 2.5622 & 2.0049 & 1.7752 & 1.6654 & 1.6171 & 1.5971 & 1.5879 & 1.5844 & 1.5827 \\
Dpn $\times 128$ & 2.5138 & 1.9650 & 1.7499 & 1.6501 & 1.6102 & 1.5917 & 1.5851 & \cellcolor{gray!30} 1.5826 & 1.5822 \\ 
GaS $\times 128$ & \cellcolor{gray!30} 2.5137 & \cellcolor{gray!30} 1.9564 & \cellcolor{gray!30} 1.7450 & \cellcolor{gray!30} 1.6471 & \cellcolor{gray!30} 1.6085 & \cellcolor{gray!30} 1.5905 & \cellcolor{gray!30} 1.5848 & \cellcolor{gray!30} 1.5826 & \cellcolor{gray!30} 1.5821 \\ \midrule
\multicolumn{10}{c}{\textbf{Gaussian, n = 50}}                                                                                                                                           \\ \midrule
Eqt $\times 128$ & 2.1551 & 1.6405 & 1.4487 & 1.3725 & 1.3464 & 1.3368 & 1.3326 & 1.3318 & 1.3312 \\
Dpn $\times 128$ & 2.0672 & 1.5964 & 1.4250 & 1.3612 & 1.3406 & 1.3340 & \cellcolor{gray!30} 1.3318 & \cellcolor{gray!30} 1.3311 & \cellcolor{gray!30} 1.3310 \\ 
GaS $\times 128$ & \cellcolor{gray!30} 2.0535 & \cellcolor{gray!30} 1.5897 & \cellcolor{gray!30} 1.4216 & \cellcolor{gray!30} 1.3597 & \cellcolor{gray!30} 1.3404 & \cellcolor{gray!30} 1.3338 & \cellcolor{gray!30} 1.3318 & \cellcolor{gray!30} 1.3311 & \cellcolor{gray!30} 1.3310 \\ \midrule
\multicolumn{10}{c}{\textbf{Explosion, n = 50}}                                                                                                                                           \\ \midrule
Eqt $\times 128$ & 2.5678 & 2.0384 & 1.8003 & 1.7099 & 1.6628 & 1.6449 & 1.6366 & 1.6351 & 1.6344 \\
Dpn $\times 128$ & 2.5231 & 2.0057 & 1.7867 & 1.6970 & \cellcolor{gray!30} 1.6538 & 1.6402 & 1.6355 & 1.6343 & \cellcolor{gray!30} 1.6341 \\ 
GaS $\times 128$ & \cellcolor{gray!30} 2.5177 & \cellcolor{gray!30} 2.0023 & \cellcolor{gray!30} 1.7846 & \cellcolor{gray!30} 1.6962 & 1.6550 & \cellcolor{gray!30} 1.6400 & \cellcolor{gray!30} 1.6354 & \cellcolor{gray!30} 1.6342 & \cellcolor{gray!30} 1.6341 \\ \bottomrule

\multicolumn{10}{c}{\textbf{Rotation, n = 100}}                                                                                                                                           \\ \midrule
Eqt $\times 128$ & 3.5046 & 2.5504 & 2.1087 & 1.8745 & 1.7562 & 1.6927 & 1.6635 & 1.6468 & 1.6392 \\
Dpn $\times 128$ & 3.3594 & 2.4615 & 2.0548 & 1.8440 & 1.7411 & 1.6858 & 1.6571 & 1.6439 & 1.6382 \\ 
GaS $\times 128$ & \cellcolor{gray!30} 3.3112 & \cellcolor{gray!30} 2.4328 & \cellcolor{gray!30} 2.0340 & \cellcolor{gray!30} 1.8378 & \cellcolor{gray!30} 1.7365 & \cellcolor{gray!30} 1.6854 & \cellcolor{gray!30} 1.6551 & \cellcolor{gray!30} 1.6432 & \cellcolor{gray!30} 1.6369 \\ \midrule
\multicolumn{10}{c}{\textbf{Gaussian, n = 100}}                                                                                                                                           \\ \midrule
Eqt $\times 128$ & 3.1191 & 2.2778 & 1.8371 & 1.6062 & 1.4987 & 1.4519 & 1.4315 & 1.4243 & 1.4230 \\
Dpn $\times 128$ & 2.9806 & 2.1438 & 1.7532 & 1.5704 & 1.4856 & 1.4478 & \cellcolor{gray!30} 1.4296 & \cellcolor{gray!30} 1.4235 & \cellcolor{gray!30} 1.4223 \\ 
GaS $\times 128$ & \cellcolor{gray!30} 2.9184 & \cellcolor{gray!30} 2.0929 & \cellcolor{gray!30} 1.7326 & \cellcolor{gray!30} 1.5657 & \cellcolor{gray!30} 1.4833 & \cellcolor{gray!30} 1.4474 & 1.4309 & 1.4240 & 1.4225 \\ \midrule
\multicolumn{10}{c}{\textbf{Explosion, n = 100}}                                                                                                                                           \\ \midrule
Eqt $\times 128$ & 3.4369 & 2.5637 & 2.1442 & 1.9181 & 1.7997 & 1.7446 & 1.7180 & 1.7061 & 1.7007 \\
Dpn $\times 128$ & 3.2976 & 2.4687 & 2.0818 & 1.8938 & 1.7908 & \cellcolor{gray!30} 1.7393 & \cellcolor{gray!30} 1.7153 & \cellcolor{gray!30} 1.7034 & \cellcolor{gray!30} 1.6991 \\ 
GaS $\times 128$ & \cellcolor{gray!30} 3.2806 & \cellcolor{gray!30} 2.4472 & \cellcolor{gray!30} 2.0756 & \cellcolor{gray!30} 1.8908 & \cellcolor{gray!30} 1.7898 & \cellcolor{gray!30} 1.7393 & 1.7154 & 1.7043 & \cellcolor{gray!30} 1.6991 \\ \bottomrule
\end{tabular*}%
}
\end{table*}

\begin{figure}[!htbp]
    \centering
    \includegraphics[width=\linewidth]{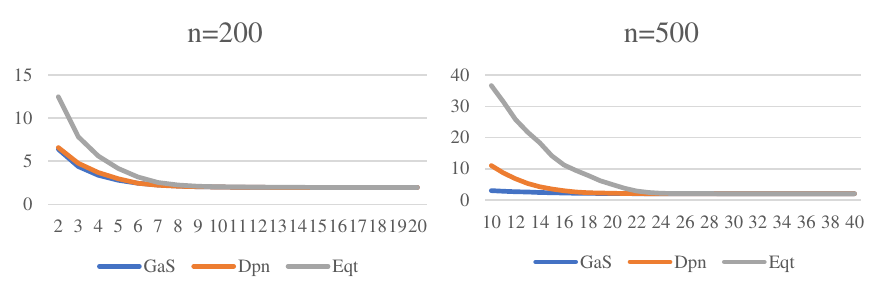}
    \vspace{-24pt}
    \caption{Performance of Finetuned Models on Larger Problem Instances. X-axis: number of agents; Y-axis: min-max cost.}
    \label{fig:trans_large}
\end{figure}

\textbf{Results for RQ2.2 \& RQ2.3}: Following the fine-tuning setup in Dpn, we fine-tune our model on instances with 200 nodes and 10–20 agents, as well as 500 nodes and 30–50 agents. Evaluation is then conducted on scenarios with 200 nodes and 2–20 agents, and 500 nodes and 10–40 agents. The results appear in Figure~\ref{fig:trans_large}.
As shown in the figure, the min-max cost remains unchanged once the number of agents exceeds a certain threshold. For the 200-node and 500-node scenarios, GaS reaches this threshold at $m = 16$ and $m = 33$, with corresponding costs of $1.9628$ and $2.0061$, respectively. Beyond these points, adding more agents yields no further improvement.
This also suggests that the number of agents used in existing evaluations for larger-scale scenarios may be suboptimal, as it often falls within a saturated range, where the optimal solution is effectively bounded by twice the distance between the depot and the farthest city node, thus limiting the potential difficulty of the instance.
For larger-scale problems (e.g., $n=1000, m \geq 50$; $n=2000, m \geq 100$; and $n=5000, m \geq 300$)—all of which are configurations used in Eqt for performance evaluation—we empirically verify that the model fine-tuned on the $n=500$ setting consistently yields optimal solutions. Therefore, we do not include additional comparisons for these configurations.
Consistent with the findings in RQ1, the more challenging aspects for current learning-based frameworks tend to emerge when the problem characteristics become more similar to those of the classical TSP. From the portion of the figure where the number of agents decreases, we observe that when a scale shift occurs, GaS demonstrates better transferability.

\subsection{Ablation study}

We conduct an ablation study to assess the contribution of each component in the GaS framework. The results are summarized in Table~\ref{tab:ablation_study}. Specifically, we examine the effects of removing the LSTM-based decoder and the optimal splitting component, denoted as w/o LSTM and w/o optimal splitting, respectively. Without the LSTM, the context embedding is processed by a multilayer perceptron (MLP) and passed directly to the multi-head attention layer. When the optimal splitting module is removed, the action of returning to the depot is no longer masked, and the decision of when to return is entirely governed by the RL policy. The results demonstrate that incorporating the LSTM effectively mitigates partial observability and enhances solution quality. Moreover, the inclusion of optimal splitting allows the RL policy to concentrate on navigation-related decisions rather than the splitting strategy, leading to substantial performance improvements, particularly in scenarios with a small number of agents.

\begin{table}[]
\centering
\caption{Ablation study on the lstm-based decoder and optimal splitting.}
\label{tab:ablation_study}
\begin{tabular}{ccccc} \toprule
\multicolumn{5}{c}{\textbf{mTSP, n=50}} \\
\midrule
m=                    & 2 & 3      & 5      & 7     \\
\midrule
w/o LSTM              & 3.2272 & 2.4862 & 2.0489 & 1.9482 \\
w/o optimal splitting & 4.3563 & 3.2291 & 2.0608 & 1.9470 \\
GaS $\times 128$ & \textbf{3.1918} & \textbf{2.4598} & \textbf{2.0301} & \textbf{1.9440} \\
\bottomrule
\end{tabular}
\end{table}

\section{Related Work}

Combinatorial optimization problems are a category of problems that focus on optimization within discrete spaces \cite{Mazyavkina2020ReinforcementLF}. 
One of the most well-known combinatorial optimization problems is the Traveling Salesman Problem \cite{Papadimitriou1977TheET} (TSP),  with many learning-based solvers \cite{Seiler2023UsingRL,Joshi2019AnEG,Holbein2023ReinforcementLO}.
As an important generalization, the $m^3$-TSP introduces additional complexity by involving multiple salesmen.
We summarize existing RL-based approaches for solving the $m^3$-TSP as follows.
NCE \cite{Kim2023LearningTC} learns to iteratively refine an initial solution through cross-exchange operations. Although this method can achieve high-quality solutions given sufficient time, it incurs a relatively high computational cost.
DisPN \cite{Hu2020ARL} develops an offline policy for assigning cities to salesmen using graph networks and reinforcement learning, while relying on traditional solvers to construct tours for each salesman.
DAN \cite{Cao2021DANDA} employs a decentralized attention-based model to learn collaborative policies for online tour construction.
ScheduleNet \cite{Park2023LearnTS} leverages type-aware graph attention to learn assignments between salesmen and cities, enabling multiple salesmen to plan tours concurrently in an online setting.
SplitNet \cite{Liang2023SplitNetAR}, which is most similar to our approach, uses classical heuristics to generate a city sequence and learns a splitting strategy to divide it into individual tours. In contrast, we adopt an optimal splitting algorithm to further improve performance.
AMARL \cite{Gao2023AMARLAA} introduces a gated transformer-based state representation to coordinate multiple agents in solving the $m^3$-TSP via RL. However, the multi-agent setting increases training complexity, while our method uses a fully serial path generation strategy, simplifying learning.
Equity-Transformer (Eqt) \cite{Son2023EquityTransformerSN} sequentially generates tours for all salesmen. Although it also uses a sequential approach, it requires learning a “return-to-depot” action, which we avoid by decoupling path generation and partitioning, thereby reducing learning difficulty.
Dpn \cite{Dpn} proposes an end-to-end model that jointly learns navigation and partition. In contrast, our method focuses solely on learning the navigation component, leaving the partitioning to be handled optimally by the optimal splitting algorithm.
UDC \cite{zheng2024udc} also follows a similar divide-and-conquer paradigm for solving combinatorial problems. While we adopt an algorithmic framework augmented with learning, they employ a fully learning-based framework.

\section{Conclusion}\label{sec:conclusion}

We introduce the Generate and Split (GaS) framework for solving the $m^3$-TSP. 
This framework comprises an LSTM-based model combined with RL for the path generator and a deterministic, approximately optimal splitting algorithm.
To train the path generator alongside the optimal splitting algorithm, we propose an LSTM-enhanced decoder that takes into account the observation history to mitigate the partial observability issue caused by delayed splitting.
Experimental results demonstrate that the GaS framework outperforms existing learning-based methods on both solution quality and transferability.

\textbf{Limitations and future work}: Unlike fully learning-based methods, the GaS framework combines algorithmic design with reinforcement learning, which may limit its immediate generality. Nevertheless, combining algorithmic priors with end-to-end learning provides a promising approach, and we plan to apply it to other combinatorial optimization problems in our future work.

\newpage



\begin{ack}
This work is supported by the National Key Research and Development Project of China No. 2023YFC3107100, NSFC No. 62172203, the Collaborative Innovation Center of Novel Software Technology and Industrialization.
\end{ack}


\bibliography{ecai-sample-and-instructions}

\begin{thebibliography}{37}
\providecommand{\natexlab}[1]{#1}
\providecommand{\url}[1]{\texttt{#1}}
\expandafter\ifx\csname urlstyle\endcsname\relax
  \providecommand{\doi}[1]{doi: #1}\else
  \providecommand{\doi}{doi: \begingroup \urlstyle{rm}\Url}\fi

\bibitem[Bogyrbayeva et~al.(2023)Bogyrbayeva, Yoon, Ko, Lim, Yun, and
  Kwon]{bogyrbayeva2022deepreinforcementlearningapproach}
A.~Bogyrbayeva, T.~Yoon, H.~Ko, S.~Lim, H.~Yun, and C.~Kwon.
\newblock A deep reinforcement learning approach for solving the traveling
  salesman problem with drone.
\newblock \emph{Transportation Research Part C: Emerging Technologies},
  148:\penalty0 103981, 2023.

\bibitem[Cao et~al.(2021)Cao, Sun, and Sartoretti]{Cao2021DANDA}
Y.~Cao, Z.~Sun, and G.~Sartoretti.
\newblock Dan: Decentralized attention-based neural network to solve the minmax
  multiple traveling salesman problem.
\newblock \emph{arXiv preprint arXiv:2109.04205}, 2021.

\bibitem[Chakraa et~al.(2023)Chakraa, Leclercq, Gu{\'e}rin, and
  Lefebvre]{Chakraa2023AMM}
H.~Chakraa, E.~Leclercq, F.~Gu{\'e}rin, and D.~Lefebvre.
\newblock A multi-robot mission planner by means of beam search approach and
  2-opt local search.
\newblock In \emph{2023 9th International Conference on Control, Decision and
  Information Technologies (CoDIT)}, pages 1--6. IEEE, 2023.

\bibitem[Cplex(2009)]{cplex2009v12}
I.~I. Cplex.
\newblock V12. 1: User’s manual for cplex.
\newblock \emph{International Business Machines Corporation}, 46\penalty0
  (53):\penalty0 157, 2009.

\bibitem[Fran{\c{c}}a et~al.(1995)Fran{\c{c}}a, Gendreau, Laporte, and
  M{\"u}ller]{Frana1995TheMS}
P.~M. Fran{\c{c}}a, M.~Gendreau, G.~Laporte, and F.~M. M{\"u}ller.
\newblock The m-traveling salesman problem with minmax objective.
\newblock \emph{Transportation Science}, 29\penalty0 (3):\penalty0 267--275,
  1995.

\bibitem[Gan and Liu(2022)]{Gan2022ForestFF}
F.~Gan and D.~Liu.
\newblock Forest fire fast response method based on optimized uav algorithm.
\newblock In \emph{2022 7th International Conference on Intelligent Computing
  and Signal Processing (ICSP)}, pages 47--51. IEEE, 2022.

\bibitem[Gao et~al.(2023)Gao, Zhou, Xu, Lan, and Xiao]{Gao2023AMARLAA}
H.~Gao, X.~Zhou, X.~Xu, Y.~Lan, and Y.~Xiao.
\newblock Amarl: An attention-based multiagent reinforcement learning approach
  to the min-max multiple traveling salesmen problem.
\newblock \emph{IEEE Transactions on Neural Networks and Learning Systems},
  2023.

\bibitem[Hausknecht and Stone(2015)]{Hausknecht2015DeepRQ}
M.~J. Hausknecht and P.~Stone.
\newblock Deep recurrent q-learning for partially observable mdps.
\newblock In \emph{AAAI fall symposia}, volume~45, page 141, 2015.

\bibitem[Helsgaun(2017)]{Helsgaun2017AnEO}
K.~Helsgaun.
\newblock An extension of the lin-kernighan-helsgaun tsp solver for constrained
  traveling salesman and vehicle routing problems.
\newblock \emph{Roskilde: Roskilde University}, 12:\penalty0 966--980, 2017.

\bibitem[Hochreiter and Schmidhuber(1997)]{Hochreiter1997LongSM}
S.~Hochreiter and J.~Schmidhuber.
\newblock Long short-term memory.
\newblock \emph{Neural computation}, 9\penalty0 (8):\penalty0 1735--1780, 1997.

\bibitem[Holbein and Schmid(2023)]{Holbein2023ReinforcementLO}
L.~Holbein and Y.~Schmid.
\newblock Reinforcement learning of tsp heuristics with message passing neural
  networks.
\newblock 2023.

\bibitem[Hu et~al.(2020)Hu, Yao, and Lee]{Hu2020ARL}
Y.~Hu, Y.~Yao, and W.~S. Lee.
\newblock A reinforcement learning approach for optimizing multiple traveling
  salesman problems over graphs.
\newblock \emph{Knowledge-Based Systems}, 204:\penalty0 106244, 2020.

\bibitem[Iklassov et~al.(2024)Iklassov, Sobirov, Solozabal, and
  Tak{\'a}{\v{c}}]{pmlr-v222-iklassov24a}
Z.~Iklassov, I.~Sobirov, R.~Solozabal, and M.~Tak{\'a}{\v{c}}.
\newblock Reinforcement learning for solving stochastic vehicle routing
  problem.
\newblock In \emph{Asian Conference on Machine Learning}, pages 502--517. PMLR,
  2024.

\bibitem[Joshi et~al.(2019)Joshi, Laurent, and Bresson]{Joshi2019AnEG}
C.~K. Joshi, T.~Laurent, and X.~Bresson.
\newblock An efficient graph convolutional network technique for the travelling
  salesman problem.
\newblock \emph{arXiv preprint arXiv:1906.01227}, 2019.

\bibitem[Kaelbling et~al.(1998)Kaelbling, Littman, and
  Cassandra]{Kaelbling1998PlanningAA}
L.~P. Kaelbling, M.~L. Littman, and A.~R. Cassandra.
\newblock Planning and acting in partially observable stochastic domains.
\newblock \emph{Artificial intelligence}, 101\penalty0 (1-2):\penalty0 99--134,
  1998.

\bibitem[Kim et~al.(2023)Kim, Park, and Park]{Kim2023LearningTC}
M.~Kim, J.~Park, and J.~Park.
\newblock Learning to cross exchange to solve min-max vehicle routing problems.
\newblock In \emph{The Eleventh International Conference on Learning
  Representations}, 2023.

\bibitem[Kool et~al.(2019)Kool, van Hoof, and Welling]{Kool2018AttentionLT}
W.~Kool, H.~van Hoof, and M.~Welling.
\newblock Attention, learn to solve routing problems!, 2019.
\newblock URL \url{https://arxiv.org/abs/1803.08475}.

\bibitem[Kwon et~al.(2020)Kwon, Choo, Kim, Yoon, Gwon, and Min]{Kwon2020POMOPO}
Y.-D. Kwon, J.~Choo, B.~Kim, I.~Yoon, Y.~Gwon, and S.~Min.
\newblock Pomo: Policy optimization with multiple optima for reinforcement
  learning.
\newblock \emph{Advances in Neural Information Processing Systems},
  33:\penalty0 21188--21198, 2020.

\bibitem[Li et~al.(2021)Li, Yan, and Wu]{Li2021LearningTD}
S.~Li, Z.~Yan, and C.~Wu.
\newblock Learning to delegate for large-scale vehicle routing.
\newblock \emph{Advances in Neural Information Processing Systems},
  34:\penalty0 26198--26211, 2021.

\bibitem[Liang et~al.(2023)Liang, Ma, Cao, Liu, Ni, Li, and
  Hao]{Liang2023SplitNetAR}
H.~Liang, Y.~Ma, Z.~Cao, T.~Liu, F.~Ni, Z.~Li, and J.~Hao.
\newblock Splitnet: a reinforcement learning based sequence splitting method
  for the minmax multiple travelling salesman problem.
\newblock In \emph{Proceedings of the AAAI Conference on Artificial
  Intelligence}, volume~37, pages 8720--8727, 2023.

\bibitem[Mahmoudinazlou and Kwon(2024)]{mahmoudinazlou2024hybrid}
S.~Mahmoudinazlou and C.~Kwon.
\newblock A hybrid genetic algorithm for the min--max multiple traveling
  salesman problem.
\newblock \emph{Computers \& Operations Research}, 162:\penalty0 106455, 2024.

\bibitem[Mazyavkina et~al.(2021)Mazyavkina, Sviridov, Ivanov, and
  Burnaev]{Mazyavkina2020ReinforcementLF}
N.~Mazyavkina, S.~Sviridov, S.~Ivanov, and E.~Burnaev.
\newblock Reinforcement learning for combinatorial optimization: A survey.
\newblock \emph{Computers \& Operations Research}, 134:\penalty0 105400, 2021.

\bibitem[Mesquita et~al.(2020)Mesquita, Souza, and Kaski]{NEURIPS2020_1764183e}
D.~Mesquita, A.~Souza, and S.~Kaski.
\newblock Rethinking pooling in graph neural networks.
\newblock \emph{Advances in Neural Information Processing Systems},
  33:\penalty0 2220--2231, 2020.

\bibitem[Papadimitriou(1977)]{Papadimitriou1977TheET}
C.~H. Papadimitriou.
\newblock The euclidean travelling salesman problem is np-complete.
\newblock \emph{Theoretical computer science}, 4\penalty0 (3):\penalty0
  237--244, 1977.

\bibitem[Park et~al.(2023)Park, Kwon, and Park]{Park2023LearnTS}
J.~Park, C.~Kwon, and J.~Park.
\newblock Learn to solve the min-max multiple traveling salesmen problem with
  reinforcement learning.
\newblock In \emph{AAMAS}, volume~22, pages 878--886, 2023.

\bibitem[Seiler et~al.(2023)Seiler, Rook, Heins, Preu{\ss}, Bossek, and
  Trautmann]{Seiler2023UsingRL}
M.~V. Seiler, J.~Rook, J.~Heins, O.~L. Preu{\ss}, J.~Bossek, and H.~Trautmann.
\newblock Using reinforcement learning for per-instance algorithm configuration
  on the tsp.
\newblock In \emph{2023 IEEE Symposium Series on Computational Intelligence
  (SSCI)}, pages 361--368. IEEE, 2023.

\bibitem[Son et~al.(2024)Son, Kim, Choi, Kim, and
  Park]{Son2023EquityTransformerSN}
J.~Son, M.~Kim, S.~Choi, H.~Kim, and J.~Park.
\newblock Equity-transformer: Solving np-hard min-max routing problems as
  sequential generation with equity context.
\newblock In \emph{Proceedings of the AAAI Conference on Artificial
  Intelligence}, volume~38, pages 20265--20273, 2024.

\bibitem[Sun et~al.(2019)Sun, Tang, Zheng, Zhou, and Yu]{Sun2019MultirobotPP}
R.~Sun, C.~Tang, J.~Zheng, Y.~Zhou, and S.~Yu.
\newblock Multi-robot path planning for complete coverage with genetic
  algorithms.
\newblock In \emph{Intelligent Robotics and Applications: 12th International
  Conference, ICIRA 2019, Shenyang, China, August 8--11, 2019, Proceedings,
  Part V 12}, pages 349--361. Springer, 2019.

\bibitem[Wang(2025)]{gas2025github}
W.~Wang.
\newblock Generate-and-split (gas) framework for $m^3$-tsp: Code and data.
\newblock \url{https://github.com/wenwenla/ecai-2025-mtsp-solver.git}, 2025.
\newblock Accessed: 2025-08-23.

\bibitem[Wilcoxon(1945)]{wilcoxon1945individual}
F.~Wilcoxon.
\newblock Individual comparisons by ranking methods.
\newblock \emph{Biometrics Bulletin}, 1\penalty0 (6):\penalty0 80--83, 1945.
\newblock \doi{10.2307/3001968}.

\bibitem[Williams(1992)]{Williams2004SimpleSG}
R.~J. Williams.
\newblock Simple statistical gradient-following algorithms for connectionist
  reinforcement learning.
\newblock \emph{Machine learning}, 8:\penalty0 229--256, 1992.

\bibitem[Yu et~al.(2023)Yu, Shi, Feng, Tian, Luo, and Wu]{yu2023esp}
X.~Yu, R.~Shi, P.~Feng, Y.~Tian, J.~Luo, and W.~Wu.
\newblock Esp: Exploiting symmetry prior for multi-agent reinforcement
  learning.
\newblock In \emph{ECAI 2023}, pages 2946--2953. IOS Press, 2023.

\bibitem[Yu et~al.(2024{\natexlab{a}})Yu, Shi, Feng, Tian, Li, Liao, and
  Wu]{yu2024leveraging}
X.~Yu, R.~Shi, P.~Feng, Y.~Tian, S.~Li, S.~Liao, and W.~Wu.
\newblock Leveraging partial symmetry for multi-agent reinforcement learning.
\newblock In \emph{Proceedings of the AAAI Conference on Artificial
  Intelligence}, volume~38, pages 17583--17590, 2024{\natexlab{a}}.

\bibitem[Yu et~al.(2024{\natexlab{b}})Yu, Tian, Wang, Feng, Wu, and
  Shi]{yu2024adaptaug}
X.~Yu, Y.~Tian, L.~Wang, P.~Feng, W.~Wu, and R.~Shi.
\newblock Adaptaug: Adaptive data augmentation framework for multi-agent
  reinforcement learning.
\newblock In \emph{2024 IEEE International Conference on Robotics and
  Automation (ICRA)}, pages 10814--10820. IEEE, 2024{\natexlab{b}}.

\bibitem[Zheng et~al.(2024{\natexlab{a}})Zheng, Yao, Wang, Tong, Yuan, and
  Tang]{Dpn}
Z.~Zheng, S.~Yao, Z.~Wang, X.~Tong, M.~Yuan, and K.~Tang.
\newblock Dpn: decoupling partition and navigation for neural solvers of
  min-max vehicle routing problems.
\newblock In \emph{Proceedings of the 41st International Conference on Machine
  Learning}, ICML'24. JMLR.org, 2024{\natexlab{a}}.

\bibitem[Zheng et~al.(2024{\natexlab{b}})Zheng, Zhou, Xialiang, Yuan, and
  Wang]{zheng2024udc}
Z.~Zheng, C.~Zhou, T.~Xialiang, M.~Yuan, and Z.~Wang.
\newblock Udc: A unified neural divide-and-conquer framework for large-scale
  combinatorial optimization problems.
\newblock \emph{Advances in Neural Information Processing Systems},
  37:\penalty0 6081--6125, 2024{\natexlab{b}}.

\bibitem[Zhou et~al.(2023)Zhou, Wu, Song, Cao, and Zhang]{zhou2023towards}
J.~Zhou, Y.~Wu, W.~Song, Z.~Cao, and J.~Zhang.
\newblock Towards omni-generalizable neural methods for vehicle routing
  problems.
\newblock In \emph{International Conference on Machine Learning}, pages
  42769--42789. PMLR, 2023.

\end{thebibliography}

\newpage
\appendix
\section{Technical Appendix}

\subsection{Proof of Correctness for the Greedy Check}\label{sec:proof}

\begin{proof}
Suppose $s = (1, p_1, p_2, \cdots, p_i, \cdots, p_{m-1}, n)$ are the split points, $p_m = n$, then the i-th subsequence $t_i$ is defined as 

\begin{equation}\label{eq:sub_sequence}
    (z'_{s_i}, \cdots, z'_{j}, \cdots, z'_{s_{i+1}-1}), \forall j, s_i < j < s_{i+1}-1.
\end{equation}

The tour for the i-th salesman is defined as $T_i$, by adding depot nodes to both ends of $t_i$ as follows:

\begin{equation}\label{eq:tour}
    (v_0, z'_{s_i}, \cdots, z'_{j}, \cdots, z'_{s_{i+1}-1}, v_0).    
\end{equation}

The cost of $T_i$ is defined as follows, where $d$ represents the Euclidean distance metric.

\begin{equation}\label{eq:cost}
    C(T_i) = \sum_{j=1}^{|T_{i}| - 1} d(T_{i,j}, T_{i,j+1}).   
\end{equation}

Since $d$ is a distance metric, the following inequality holds:

\begin{equation}\label{eq:distance_ineq}
    d(i, j) + d(j, k) \ge d(i, k), \forall{i, j, k}
\end{equation}

Assume the greedy check outputs the following split points:
$
s = (1, p_1, p_2, \cdots, p_{m-1}, n).
$
If the split points meet the requirements, then the ``greedy check'' finds a solution, and outputs true.

Otherwise, we need to prove that no split points exist that can meet the requirements.

Due to the failure of the greedy check, we can draw the following conclusions, in which $|s$ means condition on split points $s$. 

\begin{equation}\label{eq:cost_last_1}
    C(T_m|s) > c_m,
\end{equation}

\begin{equation}
    C(T_i|s) \le c_m, \forall 1 \le i < m.
\end{equation}

Assume there exists a valid splitting point, denoted as $\hat{s}=(1, \hat{p_1}, \hat{p}_2, \cdots, \hat{p}_{m-1}, n)$, and the following inequality holds.

\begin{equation}\label{eq:cost_last_2}
    C(T_i|\hat{s}) \le c_m, \forall 1 \le i \le m.
\end{equation}

\textbf{Stage 1}
According to equations (\ref{eq:cost_last_1}) and (\ref{eq:cost_last_2}), the following equation holds:

\begin{equation}\label{eq:last_sp_1}
    \hat{p}_{m-1} > p_{m-1}
\end{equation}

The proof is as follows:

\begin{equation}\label{eq:proof_1}
    \begin{aligned}
        & C(T_m|s) > c_m, C(T_m|\hat{s}) \le c_m \\
        \Rightarrow & C(T_m|s) > C(T_m | \hat{s}) \\
        \Rightarrow & d(v_0, z'_{p_{m-1}}) + \left( \sum_{k=p_{m-1}}^{n-2} d(z'_k, z'_{k+1})\right) + d(z'_{n-1}, v_0) > \\ &d(v_0, z'_{\hat{p}_{m-1}}) + \left(\sum_{k=\hat{p}_{m-1}}^{n-2} d(z'_k, z'_{k+1})\right) + d(z'_{n-1}, v_0)
    \end{aligned}
\end{equation}

From equation (\ref{eq:proof_1}), we can observe $p_{m-1} \ne \hat{p}_{m-1}$.
if $\hat{p}_{m-1} < p_{m-1}$, rewrite equation (\ref{eq:proof_1}) as follows:

\begin{equation}\label{eq:proof_2}
    d(v_0, z'_{p_{m-1}}) > d(v_0, z'_{\hat{p}_{m-1}}) + \sum_{k=\hat{p}_{m-1}}^{p_{m-1}-1}d(z'_{k}, z'_{k+1}).
\end{equation}

Equations (\ref{eq:proof_2}) and (\ref{eq:distance_ineq}) are contradictory, therefore equation (\ref{eq:last_sp_1}) must hold.

\textbf{Stage 2}
Next, we use mathematical induction to prove $\forall 1 \le i \le m-1, p_i \ge \hat{p}_i$.

First, $p_1 \ge \hat{p}_1$. This is guaranteed by the ``greedy check'' and equation (\ref{eq:distance_ineq}).

Then, assume $p_k \ge \hat{p}_k$. Next, we will prove that $p_{k+1} \ge \hat{p}_{k+1}$.

There are two possible cases, and we discuss each separately.

1. If $p_k \ge \hat{p}_{k+1}$, then $p_{k+1} \ge \hat{p}_{k+1}$.

2. If $\hat{p}_k \le p_k < \hat{p}_{k+1}$.

Let's use some simplified notation, $0 \to p_i \sim p_{i+1} \to 0$ means:

\begin{equation}
    d(v_0, z'_{p_i}) + \left(\sum_{j=p_i}^{p_{i+1}-2} d(z'_{j}, z'_{j+1})\right) + d(z'_{p_{i+1}-1}, v_0)
\end{equation}

From $p_k \ge \hat{p}_k$ and equation (\ref{eq:distance_ineq}), we can derive the following conclusion:

\begin{equation}\label{eq:proof_3}
    0 \to p_k \sim \hat{p}_{k+1} \to 0 \le 0 \to \hat{p}_k \sim \hat{p}_{k+1} \to 0 \le c_m
\end{equation}

Therefore, according to equation (\ref{eq:proof_3}), and due to the greedy nature, the conclusion $p_{k+1} \ge \hat{p}_{k+1}$ must hold.

Therefore, the following equation (\ref{eq:last_sp_2}) must hold.

\begin{equation}\label{eq:last_sp_2}
    p_{m-1} \ge \hat{p}_{m-1}
\end{equation}

Equations (\ref{eq:last_sp_1}) and (\ref{eq:last_sp_2}) are contradictory, therefore $\hat{s}$ does not exist.
Therefore, the ``greedy check'' correctly addresses the decision problem.
\end{proof}

\subsection{Details of the path generator}

We recommend referring to our implementation for the exact code\footnote{\url{https://github.com/wenwenla/ecai-2025-mtsp-solver/blob/main/net/tsp_lstm.py}}. 
Here, we briefly describe the architecture.

The encoder takes as input a graph represented by the 2D coordinates of all nodes with shape $(\text{batch\_size}, n_{\text{nodes}}, 2)$. 
It produces both node-level and graph-level embeddings of dimension 128. 
The architecture is organized as follows:

\begin{itemize}
    \item \textbf{Input embedding.} Separate linear layers are used to initialize three types of embeddings: (i) customer nodes, (ii) the depot node, and (iii) a scale feature defined as the ratio between the number of nodes and the number of agents.
    \item \textbf{Graph encoder.} The initialized embeddings are concatenated and passed through three stacked Graph Attention (GAT) layers, which capture spatial and structural dependencies between nodes.
    \item \textbf{Output.} The encoder outputs (i) node embeddings of size $(\text{batch\_size}, n_{\text{nodes}}, 128)$, (ii) a graph embedding obtained by mean pooling over all nodes, and (iii) the scale embedding.
\end{itemize}

The decoder is responsible for sequentially generating node selections based on the encoder outputs and contextual information. Its design integrates recurrent modeling with attention, and can be summarized as follows:

\begin{itemize}
    \item \textbf{Context embedding.} At each decoding step, the context vector is constructed by concatenating the graph embedding, the embedding of the first node, the embedding of the current node, and the scale embedding. This concatenated vector is projected into a 128-dimensional representation via a linear layer.
    \item \textbf{Sequential modeling.} The projected context is passed through an LSTM with hidden size 128 to capture temporal dependencies across decoding steps. Hidden and cell states are maintained across steps, with periodic detachment to stabilize training.
    \item \textbf{Multi-head attention.} The context embedding is transformed into query vectors, while cached node embeddings serve as keys and values. An 8-head scaled dot-product attention mechanism is applied with masking to ensure feasibility of node selection.
    \item \textbf{Projection and logits.} The attended context is projected back to a 128-dimensional vector and compared with transformed node embeddings to produce logits for all nodes. Logits are scaled with a hyperparameter $\alpha$ using the $\tanh$ function to control their range, and infeasible nodes are masked by assigning $-\infty$.
    \item \textbf{Output.} The decoder outputs a compatibility score vector over nodes at each step, which defines the probability distribution for selecting the next node.
\end{itemize}

\subsection{Training details}

The following table \ref{tab:hyper-parameters} summarizes the hyperparameters used in training our model.  
During training, we employ four Nvidia V100 GPUs and complete the process on 100-node instances in under three days. 
Fine-tuning is completed within 12 hours. 

\begin{table}[ht]
\centering
\caption{Training hyperparameters} \label{tab:hyper-parameters}
\label{tab:hyperparams}
\begin{tabular}{l c}
\hline
\textbf{Hyperparameter} & \textbf{Value} \\
\hline
Training Epochs   & 100 \\
Number of agents  & 2--10 \\
Augmentation      & 16 \\
Batch size        & 512 \\
Random seed       & 1234 \\
Optimizer         & Adam \\
Training learning rate   & $1 \times 10^{-4}$ \\
Fine-tuning learning rate & $1 \times 10^{-5}$ \\
\hline
\end{tabular}
\end{table}

\end{document}